\documentclass[10pt,twocolumn,letterpaper]{article}

\usepackage{wacv}
\usepackage{times}
\usepackage{epsfig}
\usepackage{graphicx}
\usepackage{amsmath}
\usepackage{amssymb}
\usepackage{booktabs}
\usepackage[accsupp]{axessibility} 
\usepackage{algorithm2e}
\usepackage{multicol}
\usepackage{multirow}
\usepackage{makecell}
\usepackage{diagbox}
\usepackage{caption}
\usepackage{xcolor}
\usepackage{subcaption}
\allowdisplaybreaks
\newcommand{\name}{EGN}
\def\wacvPaperID{1958} 

\wacvapplicationstrack 

\wacvfinalcopy 

\ifwacvfinal
\usepackage[breaklinks=true,bookmarks=false]{hyperref}
\else
\usepackage[pagebackref=true,breaklinks=true,colorlinks,bookmarks=false]{hyperref}
\fi

\usepackage[capitalize]{cleveref}
\crefname{section}{Sec.}{Secs.}
\Crefname{section}{Section}{Sections}
\Crefname{table}{Table}{Tables}
\crefname{table}{Tab.}{Tabs.}
\Crefname{equation}{Equation}{Equations}
\crefname{equation}{Eq.}{Eqs.}
\Crefname{algorithm}{Algorithm}{Algorithms}
\crefname{algorithm}{Alg.}{Algs.}
\begin{document}

\title{Exemplar Guided Deep Neural Network for Spatial Transcriptomics \\Analysis of Gene Expression Prediction}

\author{
Yan Yang$^{1}$ \quad Md Zakir Hossain$^{1}$ \quad Eric A Stone $^1$ \quad Shafin Rahman$^2$\\
$^1$ BDSI, Australian National University, Australia \quad $^2$ ECE, North South University, Bangladesh\\
{\tt\small \{Yan.Yang, zakir.hossain, eric.stone\}@anu.edu.au \quad  shafin.rahman@northsouth.edu}
}

\maketitle
\thispagestyle{empty}

\begin{abstract}
Spatial transcriptomics (ST) is essential for understanding diseases and developing novel treatments. It measures gene expression of each fine-grained area (i.e., different windows) in the tissue slide with low throughput. This paper proposes an Exemplar Guided Network (\name) to accurately and efficiently predict gene expression directly from each window of a tissue slide image. We apply exemplar learning to dynamically boost gene expression prediction from nearest/similar exemplars of a given tissue slide image window. Our \name~framework composes of three main components: 1) an extractor to structure a representation space for unsupervised exemplar retrievals; 2) a vision transformer (ViT) backbone to progressively extract representations of the input window; and 3) an Exemplar Bridging (EB) block to adaptively revise the intermediate ViT representations by using the nearest exemplars. Finally, we complete the gene expression prediction task with a simple attention-based prediction block. Experiments on standard benchmark datasets indicate the superiority of our approach when comparing with the past state-of-the-art (SOTA) methods.
\end{abstract}

\section{Introduction}
\label{sec:intro}
\begin{figure}[!t]
    \centering
    \includegraphics[width=0.75\linewidth]{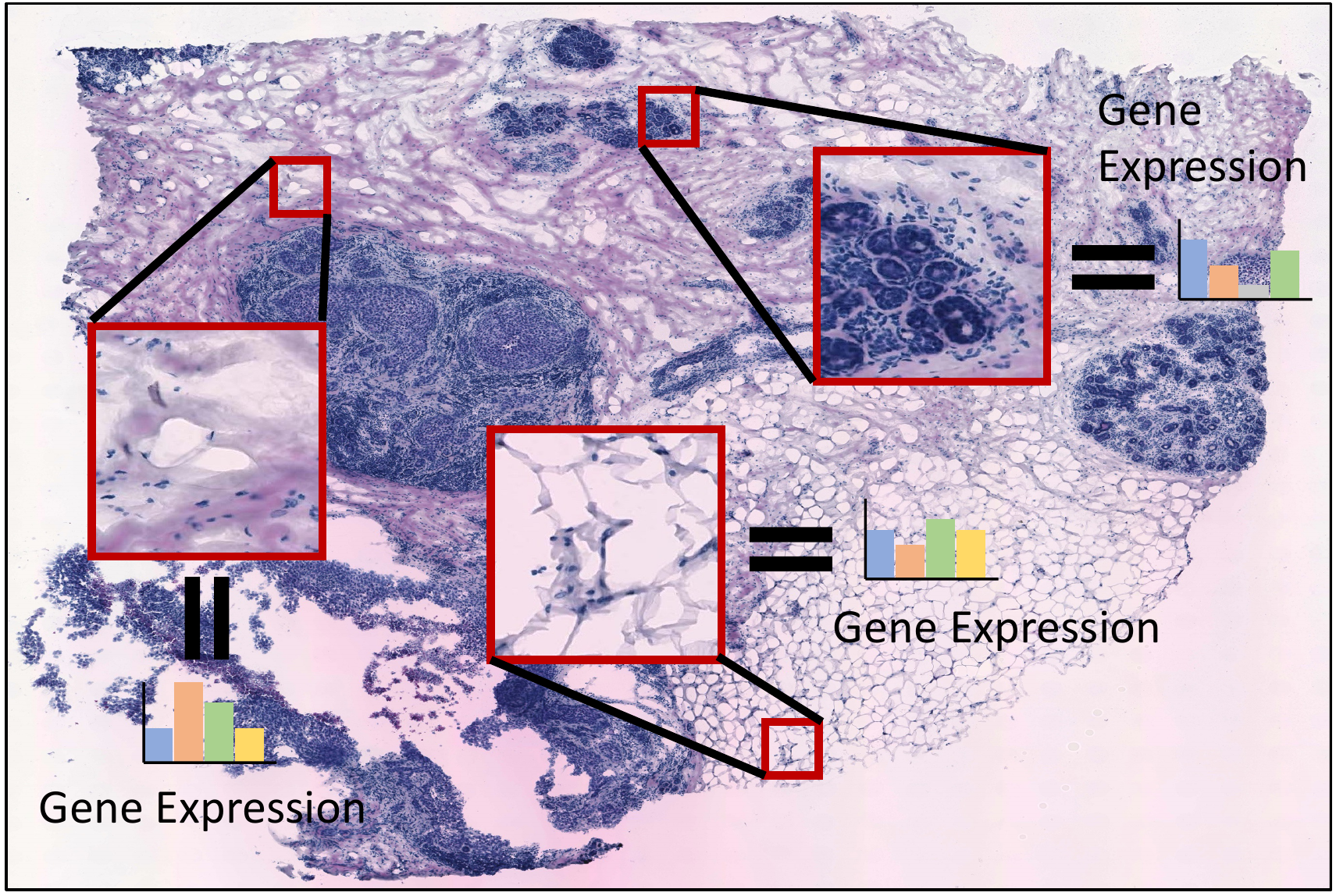}
    \caption{Overview of fields. Each fine-grained area (i.e., window) of a tissue slide image is with distinct gene expression. Here is an example, we have a tissue slide image with three windows, and each of the windows corresponds with expression of four different gene types. Our goal is to predict the gene expression of each window.
    }
    \label{fig:intro}
\end{figure}

Based on an editorial report of the Natural Methods \cite{st_nm}, Spatial Transcriptomics (ST) is the future of studying disease because of its capabilities in measuring gene expression of fine-grained areas (i.e., different windows) of tissue slides. However, ST is in low throughput due to limitations in concurrent analysis for the candidate windows \cite{st_time}. To accurately predict gene expression from each window of a  tissue slide image (\cref{fig:intro}), this paper proposes a solution named Exemplar Guided Network (\name), while allowing efficient and concurrent analysis.

In literature, previous works adopt end-to-end neural networks, namely STNet \cite{stnet} and NSL \cite{nsl}, to establish a mapping between gene expression and the slide image window. STNet is a transfer learning-based approach that finetunes a pretrained DenseNet for this task. 
On the contrary, NSL maps the color intensity of the slide image window to gene expression by a single convolution operation. Though amenable to high throughput 
because of using neural networks, their prediction performance is inferior. 

We investigate two important limitations of the existing approaches \cite{stnet,nsl}.
\textit{1)} Local feature aggregation: gene expression prediction can be considered as individually grouping the identified feature of each gene type from the slide image window. The long-range dependency among identified features is needed to reason about complex scenarios \cite{c1,c2}, as those features are generally non-uniformly distributed across the slide image window (see \cref{sec:mo} for details). STNet (i.e., a pure convolution approach) emphasizes local context during feature aggregation. It fails to bring interaction among features that are apart from each other. By experimenting with extensive SOTA architectures, we show that models with local feature aggregation achieve low performance when comparing to models with long-range dependencies.
\textit{2)} Vulnerable assumption: identifying gene expression by directly detecting the color intensity of the slide image window is vulnerable. By experimenting on standard benchmark datasets, we show this approach only works in extreme cases (for example, in the STNet dataset \cite{stnet}, tumour areas of slide image windows are usually purple, which benefits predicting tumour-related gene expression). This method shows a negative Pearson correlation coefficient (PCC) when evaluating the model with the least reliable gene expression prediction, i.e., PCC@F in \cref{tab:eva}. 

In this paper, we propose an \name~framework to address the above limitations. \name~uses ViT \cite{ViT} as a backbone and incorporates exemplar learning concepts for the gene expression prediction task. To enable unsupervised exemplar retrieval of a given slide image window, we use an extractor for defining a representation space to amount similarity between slide image windows. For brevity, we name the extractor output as a global view, which avoids confusion with the ViT representations. Note that, the representation ability of the global view is evidenced in the experiment section. Then, we have a ViT backbone to progressively construct presentations of the given slide image window under long-range dependency. Meanwhile, we have an EB block to revise intermediate ViT representations by the global views of the given slide image window, the global views of the exemplars, and the gene expression of exemplars. The global view of the given slide image window additionally serves as a prior to facilitating the long-range dependency, apart from the ViT backbone. To allow dynamic information propagation, we iteratively update the global views of the given slide image window and exemplars based on the status of the intermediate ViT representations. Semantically, the former update corresponds with `what gene expression is known ?',  and the latter corresponds with `what is gene expression the model wants to know ?'. Finally, we have an attention-based prediction block to aggregate the exemplar-revised ViT representations and the global view of the given slide image window to predict gene expression.

Our contributions are summarised below:
\begin{itemize}
    \setlength\itemsep{-0.2em}
    \item We propose an \name~framework, a ViT-based exemplar learning approach, to accurately predict gene expression from the slide image window. 
    \item We propose an EB block to revise the intermediate ViT representation by using the nearest exemplars of the given slide image window.
    \item Experiments on two standard benchmark datasets demonstrate our superiority when comparing with SOTA approaches.
\end{itemize}

\section{Related work} 
This section first review the study of gene expression prediction. Then, we summarise recent achievements of exemplar learning in both natural language processing and computer vision domains.

\noindent \textbf{Gene Expression Prediction.} 
Measuring gene expression is a basic process in developing novel treatments and monitoring human diseases \cite{enformer}. To increase process accessibility, deep learning methods have been introduced to this task. Existing methods predict the gene expression either from DNA sequences \cite{enformer} or slide images \cite{stnet,nsl,he2rna}. This paper explores the latter approach. Meanwhile, these image-based approaches are divided into two streams. First, Schmauch \textit{et al.} \cite{he2rna} employs a multi-stage method including pretrained ResNet feature extractions \cite{resnet} and a K-Means algorithm to model the Bulk RNA-Seq technique \cite{bulk}. It measures the gene expression across cells within a large predefined area that is up to $10^{5} \times 10^{5}$ pixels in a corresponding slide image \cite{he2rna}. However, this approach is ineffective for studies that require fine-grained gene expression information, such as tumour heterogeneity \cite{tumor}. Second, on the contrary, He \textit{et al.} \cite{he2rna,densenet} and Dawood \textit{et al.} \cite{nsl} design a STNet and an NSL framework to predict the gene expression for each window (i.e., fine-grained area) of a slide image. This corresponds with the ST technique \cite{st}. We model ST to predict gene expression, as this potentially solves the bulk RNA-Seq prediction task simultaneously \cite{stnet}. For example,  aggregation of gene expression predictions for each window across a slide image results in a bulk RNA-Seq prediction. 

\begin{figure*}[!t]
    \centering
    \includegraphics[width=0.9\linewidth]{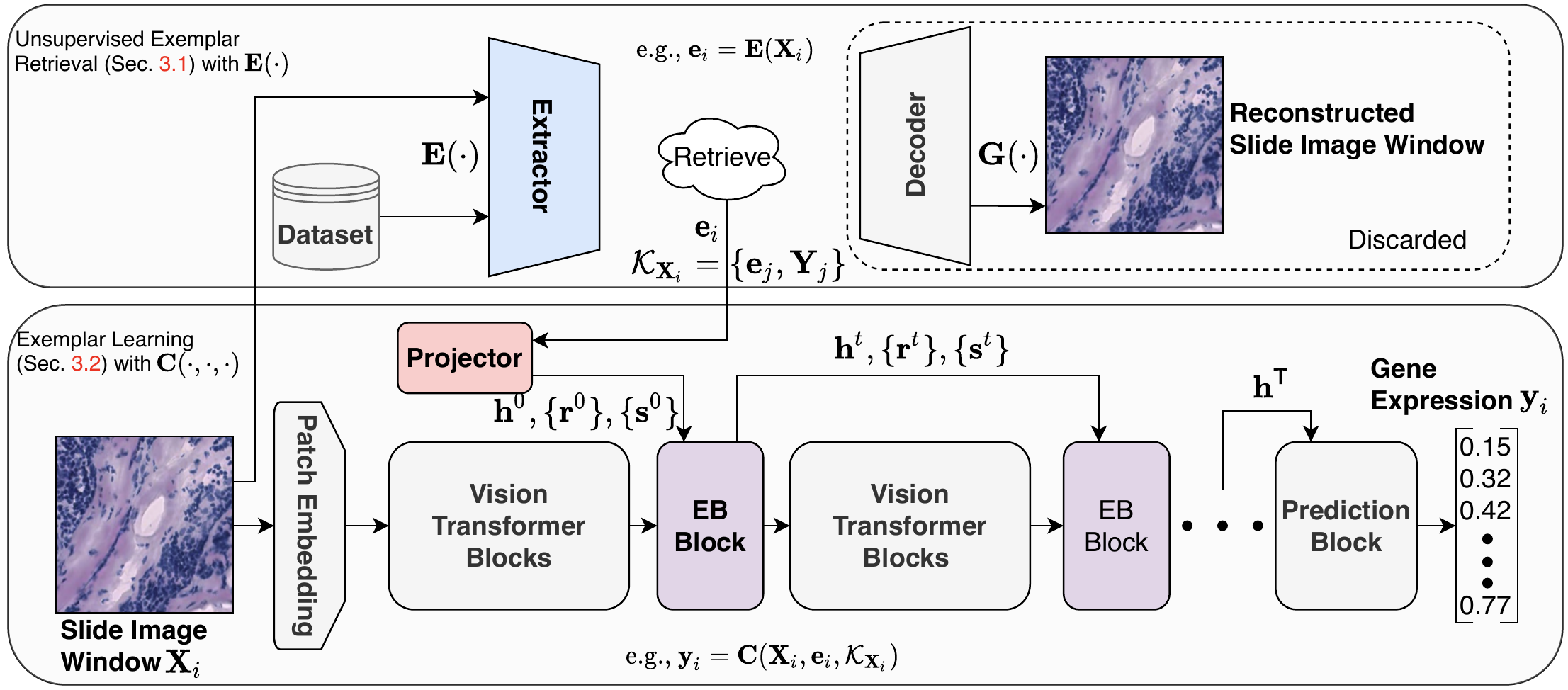}
    \caption{\name~framework. Our networks are trained in a two-stage manner. In the stage of unsupervised exemplar retrieval (\cref{sec:extractor}), we learn an extractor $\mathbf{E}(\cdot)$ and a decoder $\mathbf{G}(\cdot)$ with image reconstruction objectives. After convergence, we use the extractor $\mathbf{E}(\cdot)$ with a distance metric for unsupervised exemplar retrieval.  For example, given $\mathbf{X}_{i}$, we obtain the global view of $\mathbf{X}_{i}$, i.e., $\mathbf{e}_{i}$ and $\mathbf{e}_{i} = \mathbf{E}(\mathbf{X}_{i})$, and construct the nearest exemplar set $\mathcal{K}_{\mathbf{X}_{i}} = \{\mathbf{e}_{j},\mathbf{y}_{j}\}$, where $\mathbf{e}_{j}$ is  the exemplar global view,  and $\mathbf{y}_{j}$ is  the exemplar gene expression. 
    In the stage of exemplar learning (\cref{sec:predict}), we train a network $\mathcal{C}(\cdot,\cdot,\cdot)$ to predict gene expression $\mathbf{y}_{i}$ from $\mathbf{X}_{i}$, $\mathbf{e}_{i}$, and $\mathcal{K}_{\mathbf{X}_{i}}$. We use a vision transformer (ViT) as our backbone. The proposed EB block is interleaved with the vision transformer blocks. With a projector, we refine  $\mathbf{e}_{i}$ and $\mathcal{K}_{\mathbf{X}_{i}} = \{\mathbf{e}_{j},\mathbf{y}_{j}\}$ to $\mathbf{h}^{0}$, $\{\mathbf{r}^{0}\}$, and $\{\mathbf{s}^{0}\}$. Then, they are used by the EB block to revise the ViT patch representation and are updated to $\mathbf{h}^{t}$, $\{\mathbf{r}^{t}\}$, and $\{\mathbf{s}^{t}\}$, where $1 \leq t \leq \mathsf{T}$, and $\mathsf{T}$ is the number of layers. Finally, we have a prediction block that concatenates the refined global view $\mathbf{h}^{\mathsf{T}}$ of $\mathbf{X}_{i}$ and the attention-pooled ViT patch representation, to achieve the gene expression prediction task.
    }
    \label{fig:model}
\end{figure*}

\noindent \textbf{Exemplar Learning.} The K-nearest neighborhood classifier is the most straightforward case of exemplar learning. It classifies the input by considering the class labels of the nearest neighbors.  Exemplar learning is a composition of retrieval tasks and learning tasks \cite{clipcnn}. It has been widely employed to increase the model capability by bringing in extra knowledge of similar exemplars \cite{vqa1,vqa2,vqa3,vqa4,retro,mtr,gtm,rea,vtr,fac,fae,faf,dia}.  Patro \textit{et al.} \cite{vqa1}, Chen \textit{et al.} \cite{vqa2},  Teney \textit{et al.} \cite{vqa3}, and Farazi \textit{et al.} \cite{vqa4} refine the representation of given input with the closest match of text embedding and/or other visual embeddings to help the Visual Question Answering task. Borgeaud \textit{et al.} \cite{retro}, Wu \textit{et al.} \cite{mtr}, Urvashi \textit{et al.} \cite{vtr}, and Kelvin \textit{et al.} \cite{rea} combine the examplar learning with transformers for language generation tasks. Their key idea is to enable a more knowledgeable attention mechanism, by picking exemplars as key and value representation. Similarly, Philippe \textit{et al.} \cite{vtr} incorporates the attention mechanism with exemplar learning to bring a real-time visual object tracking framework. Other applications of exemplar learning, including fact checking, fact completion, and dialogue, could be found in \cite{fac,fae,faf,dia}. However, most of the above approach does not apply to our task because of the domain shift. This paper investigates an application of exemplar learning in gene expression prediction from slide image windows. As a result, we devise an EB block to adapt exemplar learning into our task.

\section{EGN Framework}
\label{sec:mo}

\noindent\textbf{Problem Formulation.} 
We have a dataset collection of tissue slide images, where each image contains multiple windows, and each window is annotated with gene expression. For brevity, we denote the dataset collection as pairs of a slide image window $\mathbf{X}_{i} \in \mathbb{R}^{\mathsf{3} \times \mathsf{H} \times \mathsf{W}}$and gene expression $\mathbf{y}_{i} \in \mathbb{R}^{\mathsf{M}}$, i.e., $\{(\mathbf{X}_{i}, \mathbf{y}_{i})\}_{i=1}^{\mathsf{N}}$, where $\mathsf{N}$ is the collection size, $\mathsf{H}$ and $\mathsf{W}$ are the height and width of $\mathbf{X}_{i}$, and $\mathsf{M}$ is the number of gene types. We aim to train a deep neural network model to predict $\mathbf{y}_{i}$ from $\mathbf{X}_{i}$. 

\begin{figure}[!t]
    \centering
    \includegraphics[width=0.83\linewidth]{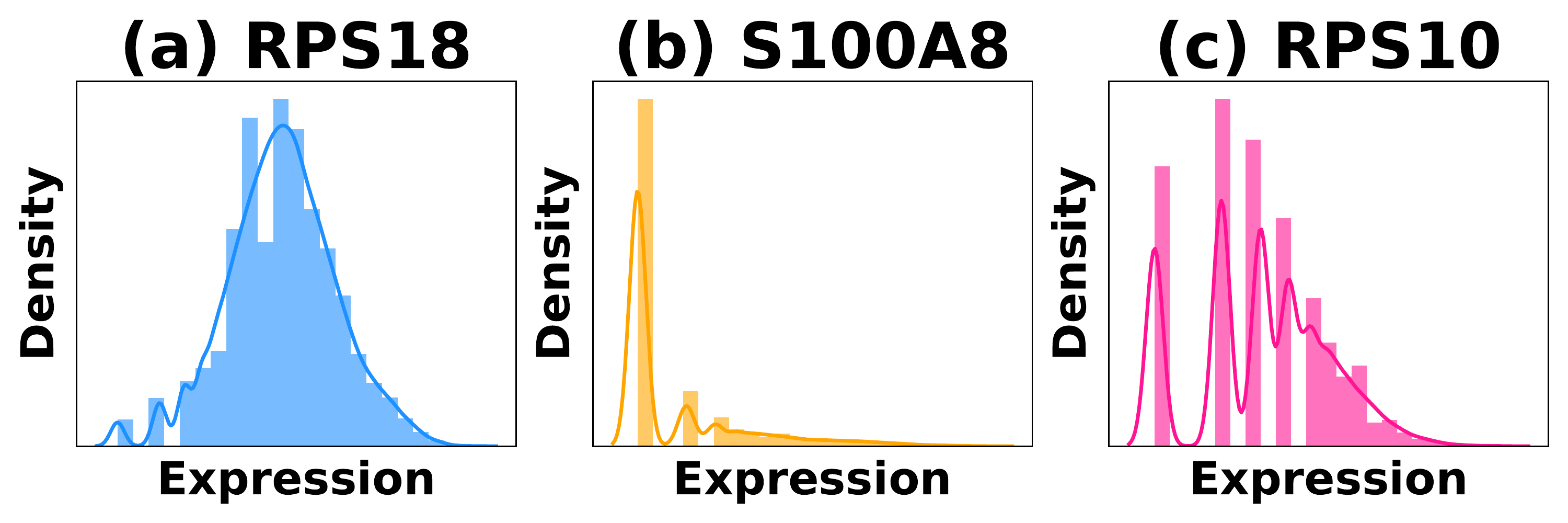}
    \vspace{-0.3em}
    \caption{Gene expression distributions of STNet dataset \cite{stnet}. Each gene expression is log-transformed. (a) is  well distributed expression of gene RPS18. (b) and (c) are long-tail distributed expression of gene S100A8 and gene RPS10.}
    
    \label{fig:dis}
\end{figure}

From the ST study, there exist two main challenges.
\textit{1)} Long-range dependency: gene expression-related features are non-uniformly distributed over the slide image window $\mathbf{X}_{i}$. Refer to Fig.~3 of \cite{stnet} for evidence. The interactions among these features are needed to group expression of the same gene type.
\textit{2)} Skewed gene expression distribution: the expression of some gene types has a skewed distribution, similar to the imbalance class distribution problem. This skewed distribution (see \cref{fig:dis} for an example) poses challenges in predicting expression of these gene types. In this paper, we attempt to mitigate them by learning from similar exemplars.

\noindent\textbf{Model Overview.} With these motivations, we have designed the \name~framework (\cref{fig:model}) containing an extractor network for exemplar retrieval and a prediction network that learns from retrieved exemplars. Networks are trained on two stages. 
\textit{1)} Unsupervised exemplar retrieval (\cref{sec:extractor}): we train an extractor $\mathbf{E}(\cdot)$ to construct the global view of the slide image window $\mathbf{X}_{i}$, i.e., $\mathbf{e}_{i} = \mathbf{E}(\mathbf{X}_{i})$ and $ \mathbf{e}_{i} \in \mathbf{R}^{\mathsf{D}}$, 
where $\mathsf{D}$ is the representation dimension. This global view $\mathbf{e}_{i}$ is used for retrieving the nearest exemplar of $\mathbf{X}_{i}$ to form a set $\mathcal{K}_{\mathbf{X}_{i}} = \{[\mathbf{e}_{j}, \mathbf{y}_{j}] \mid j \in \Upsilon_{i} \}$, where $\Upsilon_{i}$ contains indexes of nearest exemplars,  $\mathbf{e}_{j}$ is a global view of a nearest exemplar, and $\mathbf{y}_{j}$ is the paired gene expression of the exemplar.
\textit{2)} Exemplar learning (\cref{sec:predict}): we train a model $\mathbf{C}(\cdot,\cdot,\cdot)$ that maps $\mathbf{X}_{i}$, $\mathbf{e}_{i}$, and $\mathcal{K}_{\mathbf{X}_{i}}$, to $\mathbf{y}_{i}$. Our $\mathbf{C}(\cdot,\cdot,\cdot)$ is based on a ViT backbone. We bring interactions between $\mathbf{e}_{i}$ and $\mathbf{e}_{j}$ to leverage $\mathbf{y}_{j}$ with a proposed EB block. Then, for accurately predicting $\mathbf{y}_{i}$, the updated $\mathbf{e}_{i}$ is used to revise intermediate ViT representations of $\mathbf{X}_{i}$. Note that, $\mathbf{e}_{i}$ also serves as a prior to facilitating the long-range interactions. With the introduction of exemplars, our framework dynamically benefits when predicting gene expression.

\subsection{Unsupervised Exemplar Retrieval}
\label{sec:extractor}
To retrieve the exemplar of a given slide image $\mathbf{X}_{i}$ in an unsupervised manner, we propose to train an extractor (i.e., an encoder) with a decoder for image reconstructions. After convergence, we discard the decoder, and couple the extractor with a distance metric to amount the similarity between each pair of slide image windows for the exemplar retrieval.   
 
\noindent \textbf{Method.} 
The StyleGAN \cite{stylegan} applies a style code (i.e., a low dimension vector) to modulate convolution operations during image generations. As indicated by \cite{stylespace}, the style code captures both fine-grained and high-level image attributes of a training dataset in an unsupervised manner. Meanwhile, in the style code,  each scalar 
is highly disentangled, i.e., each of them tends to independently control an image attribute. This attribute independency is desired when amounting similarly between images.
Thus, we borrow the style code for our unsupervised exemplar retrieval, where the style code is a global view of the given image input. Conventionally, to obtain the style code of a given slide image window, one needs to train a styleGAN and perform a GAN inversion \cite{stylespace} separately. Instead, we jointly train an encoder/extractor $\mathbf{E}(\cdot)$ (which replaces the GAN inversion) and a StyleGAN generator $\mathbf{G}(\cdot)$, with an image reconstruction objective. Afterwards, we can directly obtain the style code by a simple forward pass in the extractor $\mathbf{E}(\cdot)$.

\noindent \textbf{Objective.} We employ a least absolute deviation loss $\mathcal{L}_{1}$, a LPIPS loss $\mathcal{L}_{\text{LPIPS}}$, and a discriminator loss $\mathcal{L}_{\mathbf{F}}$ to optimize the image reconstruction, where $\mathbf{F}(\cdot)$ is a discriminator. Note that, the reconstruction indirectly impacts the representation ability of the style code. $\mathcal{L}_{1}$ constrains pixel-wise correspondence, while $\mathcal{L}_{\text{LPIPS}}$ and $\mathcal{L}_{\mathbf{F}}$ promotes a better reconstruction fidelity. We have
\begin{align*}
    &\mathcal{L}_{1} = \lvert \mathbf{X}_{i} - \mathbf{G}(\mathbf{E}(\mathbf{X}_{i})) \rvert, \\
    &\mathcal{L}_{\text{LPIPS}} = \lVert \phi(\mathbf{X}_{i}) - \phi(\mathbf{G}(\mathbf{E}(\mathbf{X}_{i}))) \rVert_{2}, \\
    &\mathbf{L}_{\mathbf{F}} = u\Big(\mathbf{F}\big(\mathbf{X}_{i}\big)\Big) + u\Big(-\mathbf{F}\big(\mathbf{G}(\mathbf{E}(\mathbf{X}_{i}))\big)\Big),
\end{align*}
where $\phi(\cdot)$ is a pretrained LPIPS network \cite{lpips}, and $u(\cdot)$ is a Softplus function \cite{stylegan}. Let $\mathbf{X}$ be a distribution of slide image windows. The overall objective is 
\begin{equation*}
    \mathcal{L}_{\mathbf{E}} = \min_{\mathbf{G},\mathbf{E}} \max_{\mathbf{F}} ~ \mathbb{E}_{\mathbf{X}_{i} \sim \mathbf{X}} \Big[\mathcal{L}_{1} + \mathcal{L}_{\text{LPIPS}} + \mathcal{L}_{\mathbf{F}} \big ].
\end{equation*}
In a later section, we show $\mathbf{E}(\cdot)$ effectively encodes the gene expression-related features, by combining a pretrained $\mathbf{E}(\cdot)$ with a linear layer for gene expression prediction. 

\begin{figure}[!t]
    \centering
    \includegraphics[width=0.83\linewidth]{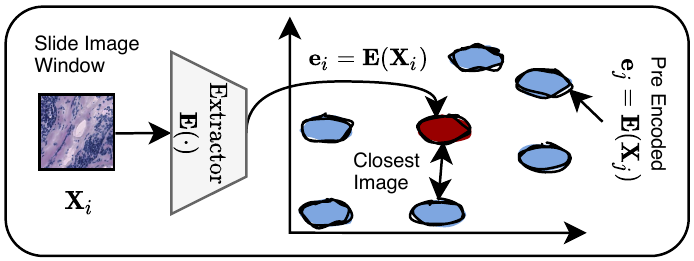}
    \caption{Overview of the exemplar retrieval. For example, blue cycles are pre encoded global views.  Given $\mathbf{X}_{i}$, we extract $\mathbf{e}_{i}$ with $\mathbf{E}(\cdot)$ and retrieve the nearest exemplars.
    }
    \label{fig:exp}
\end{figure}

\noindent\textbf{Exemplar Retrieval.} After convergence, we use the representation space of the extractor $\mathbf{E}(\cdot)$ for the exemplar retrieval. Note that, we call the extractor output, the style code, as a global view. The slide image window $\mathbf{X}_{i}$ is encoded into a global view, by $\mathbf{e}_{i} = \mathbf{E}(\mathbf{X}_{i})$. Then, we measure the similarity between each given pair of the slide image, for example, $\mathbf{X}_{i}$ and $\mathbf{X}_{j}$, with the Euclidean distance $\mathcal{L}_{2}$, for constructing the nearest exemplar set $\mathcal{K}_{\mathbf{X}_{i}}$. We empirically verify the optimal size of $\mathcal{K}_{\mathbf{X}_{i}}$ in \cref{sec:ab}. To generalize the model performance, we restrict that the candidate image pairs are from different patients. The overall process is presented in \cref{fig:exp}. In the experiment section, we compare the proposed exemplar retrieval with alternative retrieval approaches.

\subsection{Exemplar Learning}
\label{sec:predict}
Our model $\mathbf{C}(\cdot,\cdot,\cdot)$ is composed of a projector, a backbone, an EB block, and a prediction block. It maps a slide image window $\mathbf{X}_{i}$, a corresponding global view $\mathbf{e}_{i}$, and a retrieved nearest exemplar set $\mathcal{K}_{\mathbf{X}_{i}}$ to gene expression $\mathbf{y}_{i}$, i.e., $\mathbf{y}_{i} = \mathbf{C}(\mathbf{X}_{i}, \mathbf{e}_{i}, \mathcal{K}_{\mathbf{X}_{i}})$. Our model brings interactions between $\mathcal{K}_{\mathbf{X}_{i}}$ and $\mathbf{e}_{i}$ to progressively revise the intermediate representations of $\mathbf{X}_{i}$ in the ViT backbone.

\noindent\textbf{Projector.} The global views $\mathbf{e}_{i}$ and $\mathbf{e}_{j} \in \mathcal{K}_{\mathbf{X}_{i}}$ summarise a wide range of dataset-dependent attributes. We refine the global view to concentrate on the gene expression of interest by several multi-layer perceptrons (MLPs). Firstly, the global view $\mathbf{e}_{i}$ of the given slide image window $\mathbf{X}_{i}$ is projected by $\text{MLP}^{0}_{h}(\cdot)$. Secondly, for $\mathbf{e}_{j} \in \mathcal{K}_{\mathbf{X}_{i}}$, as the associated gene expression $\mathbf{y}_{i} \in \mathcal{K}_{\mathbf{X}_{i}}$ is available, we empower the refinement of $\mathbf{e}_{j}$ by $\mathbf{y}_{j}$. We concatenate $\mathbf{e}_{j}$ and $\mathbf{y}_{j}$ before feeding to $\text{MLP}^{0}_{r}(\cdot)$. Thirdly, with $\text{MLP}^{0}_{s}(\cdot)$, $\mathbf{y}_{j}$ is projected to the model dimension.  We have
\begin{align*}
    \mathbf{h}^{0}_{i} = \text{MLP}^{0}_{h}(\mathbf{e}_{i}), \;\; \mathbf{r}^{0}_{j} = \text{MLP}^{0}_{r}([\mathbf{e}^{j}, \mathbf{Y}{j}]), \;\;
    \mathbf{s}^{0}_{j} = \text{MLP}^{0}_{s}(\mathbf{y}_{j}),
\end{align*}
where the superscript of $\mathbf{h}^{0}_{i}$, $\mathbf{r}^{0}_{j}$, and $\mathbf{s}^{0}_{j}$ denotes that they are initial refined global view, and $[\cdot,\cdot]$ is a concatenation operator. $\text{MLP}^{0}_{h}(\cdot)$ and $\text{MLP}^{0}_{s}(\cdot)$ are two layer perceptrons with a ReLU activation function. $\text{MLP}^{0}_{r}(\cdot)$ shares the same parameters with $\text{MLP}^{0}_{h}(\cdot)$ but has an extra linear layer on top of it. 

\noindent \textbf{Backbone.} We use the ViT as our backbone \cite{ViT}. It has a patch embedding layer and vision transformer blocks.  The patch embedding layer tiles the slide image window $\mathbf{X}_{i}$ into non-overlapping patches before flattening and feeding to a linear layer. The outputs are $\mathbf{Z}^{0}=[\mathbf{z}^{0}_{l} \mid  l \in 1,\cdots, \mathsf{L}]$, where $\mathsf{L}$ is the total number of patches, and $\mathbf{Z}^{0}$ is a matrix representation of the projected patches. Each vision transformer block composes of a self-attention layer and a feedforward layer, which performs global interaction among $\mathbf{Z}^{0}$.  Assuming there are $\mathsf{T}$ layers, and let $t \in [1,\cdots, \mathsf{T}]$. We denote the $l^{th}$ patch representation at $t^{th}$ layer as $\mathbf{z}^{t}_{l}$. 
\begin{figure}[!t]
    \centering
    \includegraphics[width=\linewidth]{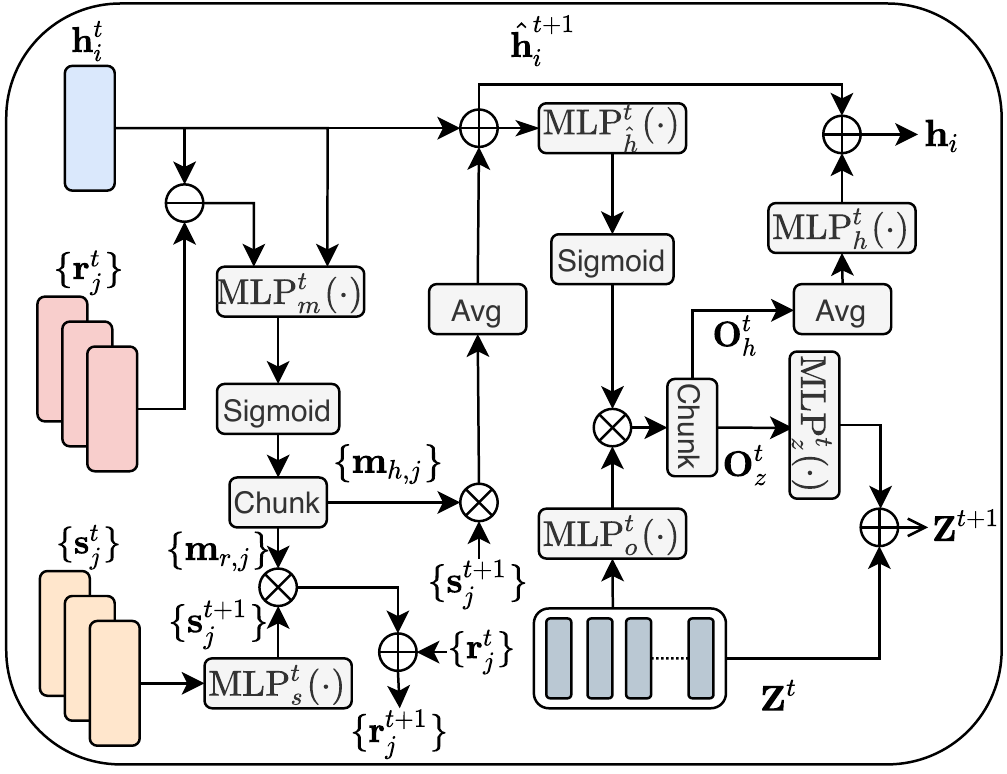}
    \caption{EB block architecture. The inputs are $\mathbf{h}_{i}$, $\{\mathbf{r}^{t}_{j} \mid j \in \Upsilon_{i} \}$, and $\{\mathbf{s}^{t}_{j} \mid j \in \Upsilon_{i} \}$, and $\mathbf{Z}^{t}$. 
    Firstly, we project $\mathbf{s}_{j}$ to $\mathbf{s}^{t+1}_{i}$ and have interactions between $\mathbf{h}_{i}$ and $\mathbf{r}_{j}$.
    This interaction is assisted by $\mathbf{s}^{t+1}_{i}$ and obtains $\hat{\mathbf{h}^{t+1}}_{i}$ and  $\mathbf{r}^{t+1}_{i}$. Secondly, $\hat{\mathbf{h}}^{t}_{i}$ is projected to the number of patches in the ViT backbone, to scale the magnitude of the ViT patch representation $\mathbf{Z}^{t}$. By their interactions,  $\hat{\mathbf{h}}^{t+1}_{i}$ and $\mathbf{Z}^{t}$ are updated to $\mathbf{h}^{t+1}_{i}$ and $\mathbf{Z}^{t+1}$. 
    }
    \label{fig:tbc}
\end{figure}
\begin{table*}[!t]
    \centering
    \caption{Quantitative gene expression prediction comparisons with SOTA methods in STNet dataset and 10xProteomic dataset. We bold the best results. We use `-' to denote unavailable results. Models are evaluated by four-fold cross-validation and three-fold cross-validation in the above datasets. Our proposed \name~framework consistently outperforms the SOTA methods in  MAE$_{\times10^{1}}$, PCC@F$_{\times10^{1}}$, PCC@S$_{\times10^{1}}$ and PCC@M$_{\times10^{1}}$ for both datasets. The CycleMLP finds the best MSE$_{\times10^{2}}$ in the 10xProteomic dataset.
    }
    \vspace{-0.7em}
    \resizebox{\linewidth}{!}{
    \begin{tabular}{c|l|cccccccc}
         \toprule
         & Methods &  STNet \cite{stnet} & NSL \cite{nsl} & ViT \cite{ViT} & CycleMLP \cite{cyclemlp} & MPViT \cite{mpViT} & Retro \cite{retro} & ViTExp & Ours \\
         \midrule
         \multirow{5}{*}{\rotatebox[origin=c]{90}{\makecell[c]{STNet\\Dataset}}} &MSE$_{\times10^{2}}$&4.52 & - &4.28 & 4.41 &4.49& 4.53 & 4.46 & \textbf{4.10}\\
         &MAE$_{\times10^{1}}$ &1.70 & - &1.67 &1.68 &1.70& 1.71& 1.69 &\textbf{1.61}\\
         &PCC@F$_{\times10^{1}}$ &0.05 & -0.71  & 0.97 &1.11 &0.91& 0.99 &0.87 &\textbf{1.51}\\
         &PCC@S$_{\times10^{1}}$&0.92&0.25  &1.86 &1.95 &1.54& 1.74 &1.72&\textbf{2.25}\\
         &PCC@M$_{\times10^{1}}$&0.93&0.11  &1.82 &1.91 &1.69&1.79 &1.74& \textbf{2.02}\\
         \midrule
         \multirow{5}{*}{\rotatebox[origin=c]{90}{\makecell[c]{10xProteomic\\Dataset}}}  &MSE$_{\times10^{2}}$& 12.40 & - & 7.54 & \textbf{4.69}&5.45 & 5.25 & 5.04 & 5.49\\
         &MAE$_{\times10^{1}}$& 2.64 & - & 2.27 & \textbf{1.55}&1.56 & 1.65 & 1.66 & \textbf{1.55}\\
         &PCC@F$_{\times10^{1}}$&1.25 &-3.73 & 5.11 & 5.88 & 6.40 & 5.46 & 5.59 & \textbf{6.78}\\
         &PCC@S$_{\times10^{1}}$&2.26 & 1.84 & 4.64 & 6.60 & 7.15 & 6.35 & 6.36 & \textbf{7.21}\\
         &PCC@M$_{\times10^{1}}$&2.15 & 0.25 & 4.90 & 6.32 & 6.84 & 6.04 & 6.00 & \textbf{7.07}\\
         \bottomrule
    \end{tabular}}
    \label{tab:eva}
\end{table*}

\noindent\textbf{EB Block.} This block is interleaved with the vision transformer blocks, which brings $\mathbf{s}^{t}_{j}$, i.e., knowledge about gene expression of the nearest exemplar, to each ViT patch representation $\mathbf{z}^{t}_{l}$. Firstly, we project $\mathbf{s}^{t}_{j}$ (i.e., a projection of gene expression of the exemplars) to $\mathbf{s}^{t+1}_{j}$. With $\mathbf{s}^{t+1}_{j}$, we have interactions between $\mathbf{h}^{t}_{i}$ and $\mathbf{r}^{t}_{j}$ (i.e., refined global views) to obtain $\hat{\mathbf{h}}^{t+1}_{i}$ and  $\mathbf{r}^{t+1}_{j}$. As $\hat{\mathbf{h}}^{t+1}_{i}$ and $\mathbf{r}^{t+1}_{j}$ are initialized from the same extractor $\mathbf{E}(\cdot)$, their interactions could serve as a bridge for the revision of the ViT patch representation. Secondly,  considering $\hat{\mathbf{h}}^{t+1}_{i}$ is initially extracted by the reconstruction based-network $\mathbf{E}(\cdot)$, $\hat{\mathbf{h}}^{t+1}_{i}$ is expected to contain spatial information about the input image distribution. With $\hat{\mathbf{h}}^{t+1}_{i}$, we revise the ViT patch representation $\mathbf{z}^{t}_{l}$, obtaining $\mathbf{h}^{t+1}_{i}$ and $\mathbf{z}^{t+1}_{l}$. This revision process brings interactions between $\hat{\mathbf{h}}^{t+1}_{i}$and $\mathbf{z}^{t}_{l}$ to propagate information about possible gene expression (which receives from the exemplars) and patch representation statuses. The overall process is shown in \cref{fig:tbc}.

To do so, firstly, we concatenate $\mathbf{h}^{t}_{i}$ and the difference $\mathbf{h}^{t}_{i} - \mathbf{r}^{t}_{j}$ to bring interactions between $\mathbf{h}^{t}_{i}$ and $\mathbf{r}^{t}_{j}$, with $\text{MLP}^{t}_{m}(\cdot)$. We chunk the outputs to $\mathbf{m}_{h,j}$ and $\mathbf{m}_{r,j}$.  Then, they are fused with $\hat{\mathbf{s}}_{j}$ to retrieve gene expression from $\mathbf{s}^{t+1}_{j}$ (a projection of $\mathbf{s}^{t}_{j}$).
Semantically, $\mathbf{m}_{h,j}$ summarises `the existing knowledge of gene expression', and
$\mathbf{m}_{r,j}$ tells `the desired knowledge'. Mathematically, we have 
\begin{align*}
    &\hat{\mathbf{h}}^{t+1}_{i} = \mathbf{h}^{t}_{i} + \text{Avg}([\mathbf{m}_{h,j} \cdot \mathbf{s}^{t+1}_{j} \mid \forall j \in \Upsilon_{i}]), \\
    &\mathbf{r}^{t+1}_{i} = \mathbf{r}^{t}_{i} + \mathbf{m}_{r,j} \cdot \mathbf{s}^{t+1}_{j}, \\
    &\mathbf{m}_{h,j}, \mathbf{m}_{r,j} = \text{Chunk}(\sigma(\text{MLP}^{t}_{m} (\mathbf{h}^{t}_{i},\mathbf{h}^{t}_{i} - \mathbf{r}^{t}_{j}))),\\
    &\mathbf{s}^{t+1}_{j} = \text{MLP}^{t}_{s}(\mathbf{s}^{t}_{j})  
\end{align*}
where $\text{MLP}^{t}_{m}(\cdot)$ is a Multi-layer perceptron, the $\text{Chunk}(\cdot)$ operator equally splits the input into two outputs, and $\sigma(\cdot)$ is a Sigmoid function. To avoid notation overloading, we remain to use $\text{MLP}^{t}_{s}(\cdot)$ as a single-layer perceptron. Secondly, $\hat{\mathbf{h}}^{t+1}_{i}$ and $\mathbf{Z}^{t}$ reciprocate each other by scaling the magnitude of each patch representation $\mathbf{z}^{t}_{l} \in \mathbf{Z}^{t}$: 
\begin{align*}
    &\mathbf{Z}^{t+1} = \mathbf{Z}^{t} + \text{MLP}^{t}_{z}(\mathbf{O}^{t}_{z}), \\
    &\mathbf{h}^{t+1}_{i} = \hat{\mathbf{h}}^{t+1}_{i} + \text{MLP}^{t}_{h}(\text{Avg}(\mathbf{O}^{t}_{h})), \\
    &\mathbf{O}^{t}_{h},\mathbf{O}^{t}_{z}= \text{Chunk}(\text{MLP}^{t}_{o}(\mathbf{Z}^{t}) \cdot \sigma(\text{MLP}^{t}_{\hat{h}}(\hat{\mathbf{h}}^{t+1}_{i}))), 
\end{align*}
where $\text{MLP}^{t}_{z}(\cdot)$, $\text{MLP}^{t}_{h}(\cdot)$, $\text{MLP}^{t}_{o}(\cdot)$, and $\text{MLP}^{t}_{\hat{h}}(\cdot)$ are single-layer perceptrons. $\mathbf{O}^{t}_{h}$ and $\mathbf{O}^{t}_{z}$ are the fusion of the ViT patch representations $\mathbf{Z}^{t}$ and $\hat{\mathbf{h}}^{t+1}_{i}$. Note that, the output dimension of $\text{MLP}^{t}_{o}(\cdot)$ is the number of patches in the ViT, and each scalar of the output is used to scale a corresponding patch from $\text{MLP}^{t}_{o}(\mathbf{Z}^{t})$. By scaling the magnitudes of each ViT patch representation, we indirectly inject the knowledge about gene expression of the exemplars, while updating the refined global view $\hat{\mathbf{h}}^{t+1}_{i}$ to facilitate both interactions with the exemplars and the ViT representation in following layers. One may note that the current operation is single-head based, which can be extended to a multi-head operation by expanding the output dimension of $\text{MLP}^{t}_{o}(\cdot)$ in practice.

\noindent\textbf{Prediction Block.} Following \cite{crs}, each ViT patch representation shows different priorities toward the gene expression prediction task. We apply an attention pooling layer $\text{AttPool}(\cdot)$ to aggregate $\mathbf{z}^{\mathsf{T}}_{l} \in \mathbf{Z}^{\mathsf{T}}$. Moreover, $\mathbf{h}^{\mathsf{T}}_{i}$ is a progressively refined global view of the slide image window $\mathbf{X}_{i}$, which captures high-level attributes of $\mathbf{X}_{i}$. We concatenate $\mathbf{h}^{\mathsf{T}}_{i}$ and $\text{AttPool}(\mathbf{Z}^{\mathsf{T}})$ for the  prediction. We have
\begin{align*}
    \mathbf{y}_{i} = \text{MLP}_{g}([ \mathbf{h}^{\mathsf{T}}_{i},\text{AttPool}(\mathbf{Z}^{\mathsf{T}})]),
\end{align*}
where $\text{MLP}_{g}$ is a single-layer perception.

\noindent \textbf{Objective.} We optimize $\mathbf{C}(\cdot,\cdot,\cdot)$ with mean squared loss (i.e., the Euclidean distance) $\mathcal{L}_{2}$ and batch-wise PCC $\mathcal{L}_{ppc}$. We have
\begin{align*}
\mathcal{L}_{total} = \mathcal{L}_{2} + \mathcal{L}_{ppc}
\end{align*}

\begin{figure*}[!t]
    \centering
    \includegraphics[width=\linewidth]{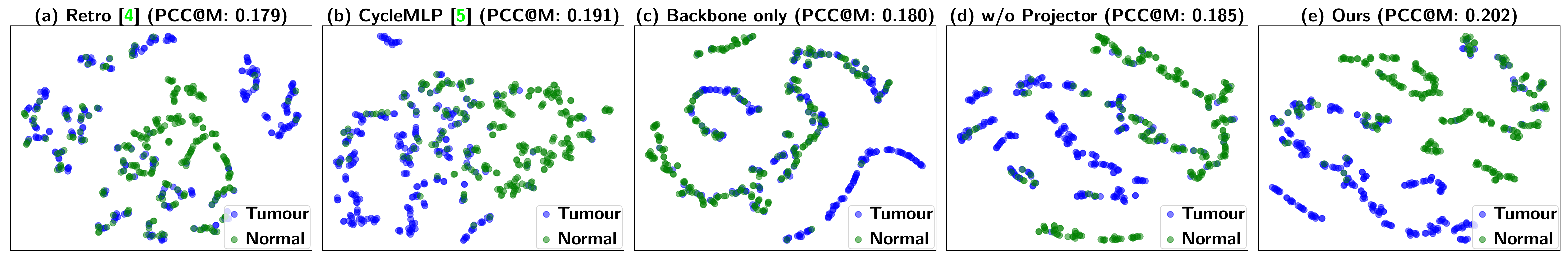}
    \caption{Quantitative evaluation of the top performed models from \cref{tab:eva} and \cref{tab:net}. We employ t-SNE \cite{tsne} for dimension reduction of model latent space. We use the extra labels (i.e., tumour and normal) from the STNet dataset for annotations.
    }
    \label{fig:vis}
\end{figure*}

\section{Experiments}
\noindent\textbf{Datasets.} We perform experiments in the publicly available STNet dataset \cite{stnet} and 10xProteomic datasets\footnote{\url{https://www.10xgenomics.com/resources/datasets}}. The STNet dataset contains roughly 30,612 pairs of the slide image window and the gene expression. This dataset covers 68 slide images from 23 patients. Following \cite{stnet}, we target predicting expression of 250 gene types that have the largest mean across the dataset. The 10xProteomic dataset has 32,032 slide image windows and gene expression pairs from 5 slide images. We select target gene types in the same way as the STNet dataset. We apply log transformation and min-max normalization to the target gene expression. Note that, our normalization method is different from \cite{stnet} (they use log transformation and a custom normalization, i.e., dividing the expression of each gene type by the sum of expression from all gene types, for each slide image window). Our normalization method allows independent analysis of gene expression prediction.

\subsection{Experimental set-up}

\noindent\textbf{Baseline methods}. We compare with extensive SOTA methods from gene expression prediction, ImageNet classification benchmarks, and exemplar learning. 
\begin{itemize}
    \setlength\itemsep{-0.2em}
    \item STNet \cite{stnet} and NSL \cite{nsl}. They are the SOTA methods in gene expression prediction. 
    \item ViT \cite{ViT}, MPViT \cite{mpViT} and CycleMLP \cite{cyclemlp}. We use the SOTA ImageNet classification methods in our task. They are strong baselines in our task.  Specifically, we use ViT-B, MPViT-Base, and CycleMLP-B2. Please refer to \cite{ViT,mpViT,cyclemlp} for details.
    \item Retro \cite{retro} and ViTExp. We explore the SOTA exemplar learning methods. However, Retro is originally developed for natural language processing. We adapt it by providing the extractor output $\mathbf{e}$ as the exemplar representations. ViTExp directly concatenates the exemplar representations to the ViT patch representation. The exemplar representations are added with an embedding to be differentiated from the patch representation of the slide image window. Both Retro and ViTExp are based on the ViT-B architecture \cite{ViT}.
\end{itemize}
     
\noindent\textbf{Evaluation metrics.} We evaluate the proposed methods and alternative baselines with PCC, mean squared error (MSE), and mean absolute error (MAE). 
We use PCC@F, PCC@S, and PCC@M to denote the first quantile, median, and mean of the PCC. 
The PCC@F verifies the least performed model predictions. The PCC@S and PCC@M measure the median and mean of correlations for each gene type, given predictions and GTs for all of the slide image windows. Meanwhile, the MSE and MAE measure the sample-wise deviation between predictions and GTs of each slide image window for each gene type. Note that, for MSE and MAE, the lower value means better performance. In contrast, for PCC@F, PCC@S, and PCC@M, the higher value indicates better performance. 

\noindent\textbf{Implementation details\footnote{Codes are available at: \url{https://github.com/Yan98/EGN}}.} The architectures and optimization settings for our extractor follow \cite{s2f}. For the exemplar learning, we implement \name~by using the \textit{Pytorch} framework \cite{pytorch}. \name~is trained from scratch for 50 epochs with batch size 32. We set the learning rate to $5 \times 10^{-4}$ with a cosine annealing scheduler. Our weight decay is $1 \times 10^{-4}$.  We use a ViT backbone with the following settings: patch size 32, embedding dimension 1024, feedforward dimension 4096, attention heads 16,  and depth 8. We interleave the proposed EB block with the ViT backbone at frequency 2, where each block has 16 heads and dimension 64, namely the block has 16 unique operations with dimension 64. Our EB block uses 9 nearest exemplars. All the experiments are conducted in 2 NVIDIA Tesla P100 GPUs.

\begin{table*}[!t]
    \begin{minipage}{0.55\linewidth}
    \centering
    \includegraphics[width=\linewidth]{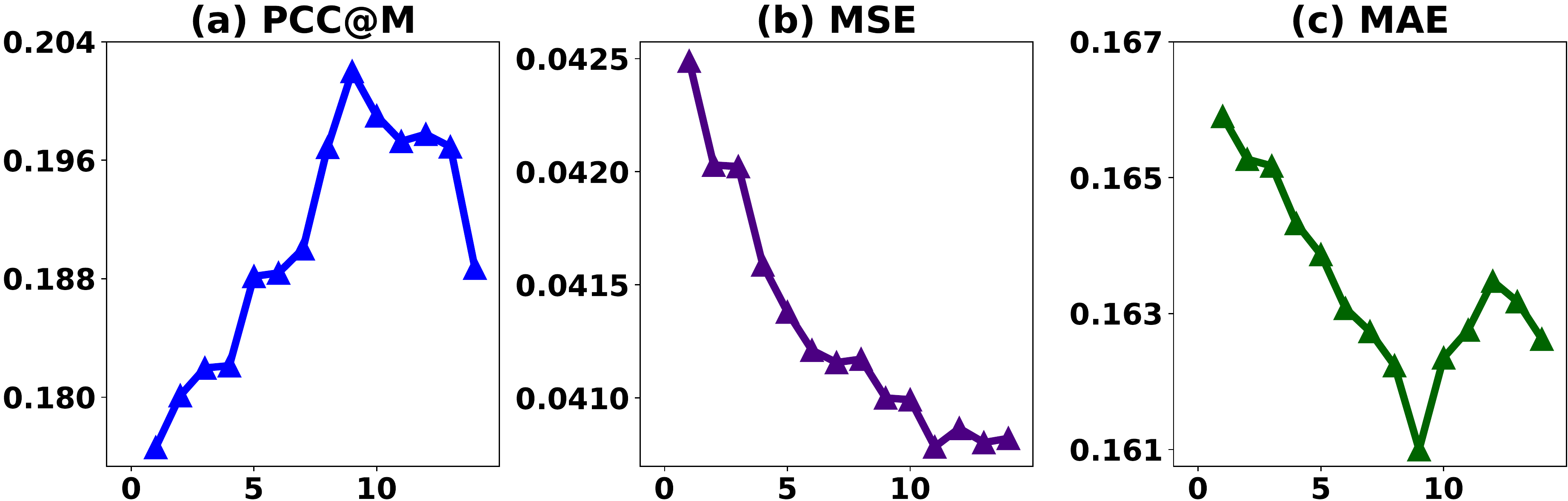}
    \captionof{figure}{Ablation study on number of exemplar used in EB block. This number is varied from one to fifteen, and we present PCC@M, MSE, and MAE in sub-figure (a), (b), and (c).
    }
    \label{fig:nmk}
    \end{minipage}
    \hspace{1em}
    \begin{minipage}{0.4\linewidth}
    \centering
    \caption{Ablation study on exemplar retrieval. We compare with representation from AlexNet and ResNet50 for exemplar retrieval.}
    \vspace{-0.7em}
    \resizebox{0.9\linewidth}{!}{
    \begin{tabular}{llccc}
       \toprule
       \multirow{2}{*}{\makecell[l]{Feature\\Space}} &  \multirow{2}{*}{\makecell[c]{Distance\\Metrics}} &  \multirow{2}{*}{MSE$_{\times10^{2}}$} &
       \multirow{2}{*}{MAE$_{\times10^{1}}$} & \multirow{2}{*}{PCC@M$_{\times10^{1}}$} \\
       & \\
       \midrule
       $\mathbf{E}(\cdot)$  & $\mathcal{L}_{2}$ & \textbf{4.10} & \textbf{1.61} & \textbf{2.02} \\
       $\mathbf{E}(\cdot)$  & $\mathcal{L}_{1}$  & 4.13 & 1.64 & 1.80 \\
       $\mathbf{E}(\cdot)$  & $cosine$ .  & 4.24 & 1.63 & 1.83 \\
       AlexNet  & LPIPS & 4.46 & 1.70 & 1.74\\
       ResNet50 & $\mathcal{L}_{2}$ & 4.12 & 1.62 & 1.89\\
       ResNet50 & $\mathcal{L}_{1}$ & 4.13 & 1.62 & 1.83\\
       ResNet50 & $cosine$ &  4.16 & 1.64 & 1.94 \\
       \bottomrule
    \end{tabular}}
    \label{tab:ret}
    \end{minipage}
\end{table*}
\subsection{Experimental results}
We compare our \name~framework with the baselines on the STNet dataset and the 10xProteomic dataset (\cref{tab:eva}). As the gene expression prediction task emphasizes capturing the relativity variation,  we bias on the PCC-related evaluation metrics, i.e., PCC@F, PCC@S, and PCC@M. Our \name~consistently achieves the SOTA performance in MAE, PCC@F, PCC@S, and PCC@M.
Our findings are as follows:
\textit{1)} It's worth noting that Retro and ViTExp use the ViT as the backbone. However, their performance is lower than the ViT. Both Retro and ViTExp use attention-based operations to directly have interactions between the ViT representations and the global views of the exemplars, when comparing with our EB block. This validates the necessity of using the global view of the input slide image window as a bridge to introduce the knowledge of exemplars to the ViT representations;
\textit{2)} CycleMLP and MPViT, the SOTA methods in the ImageNet classification task, lead to the second-best performance in the STNet dataset and 10xProteomic dataset in PCC-related evaluations. Meanwhile, CycleMLP finds the best MSE in 10xProteomic. Note that, PCC-related evaluation metrics are most important in our task. By using the exemplars, our model that uses the vanilla ViT backbone outperforms them with a reason marginal in PCC-related evaluation metrics, while overall achieving similar MSE and MAE with them;  
\textit{3)} The PCC@F of our model significantly outperforms the baseline methods. This metric evaluates the worst model capability, by calculating the first quantile of PCC across all gene types. Note that, the majority of gene types covered by the first quantile have skewed expression distributions, which are the most challenging part of the prediction task. Our method has 0.038 - 0.040 higher than the second-best method in PCC@F;
and \textit{4)} STNet and NSL fail to achieve good performance. Again, the gene expression-related feature is usually non-uniformly distributed across the slide image window input. They do not have long-range interactions to capture the expression of the same gene type from these features. Moreover, NSL even shows a negative correlation with PCC@F. This validates our claims that predicting gene expression directly with the color intensity is vulnerable, and it is only feasible in extreme cases (recall the example of tumour-related gene expression in \cref{sec:intro}).

\noindent\textbf{Quantitative Evaluation.} 
We present the latent space visualization (\cref{fig:vis}), by considering the top performed models from \cref{tab:eva} and \cref{tab:net} (see details of `Backbone only' and `w/o Projector' settings in \cref{sec:ab}). 
Note that, the `Backbone only' setting has a smaller number of parameters than a regular ViT-B. The extra labels (i.e., tumour and normal) from the STNet dataset are used for annotations. 
To enable a clean visualization, we randomly sample 256 representations of the slide image window for each label, i.e., tumour and normal. By gradually using the proposed components (from \cref{fig:vis} (c) to \cref{fig:vis} (e)), i.e., introduction exemplars to our model, our method sufficiently separates the tumour representations from the normal representations, by having an improved gene expression prediction than alternative approaches.

\subsection{Ablation study}
\label{sec:ab}
We study the capability of each model component by conducting a detailed ablation study in the STNet dataset.

\noindent\textbf{Extractor Capability.} We explore alternative approaches for retrieving exemplars (\cref{tab:ret}). We compare with representations from AlexNet \cite{alex} and ResNet50 \cite{resnet}. Note that, they are pretrained on ImageNet, and our $\mathbf{E}(\cdot)$ is learned in an unsupervised manner. We explore diverse distance matrices including LPIPS \cite{lpips}, $\mathcal{L}_{2}$, $\mathcal{L}_{1}$, and $cosine$ (i.e., cosine similarity). We have the following findings:
\textit{1)} LPIPS distance retrieved exemplars lead to bad model performance. This distance is trained based on the human perception of regular images, where these images are different from the slide image windows. Thus, it retrieves bad exemplars and damages the model performance;
\textit{2)} Retrieving exemplars with ResNet50 lead to decent performance, though the pretrained ResNet50 lacks gene expression-related knowledge. This retrieval method complements the knowledge about image textures learned from the ImageNet for our task, which enhances the versatility of our model representation and verifies our model robustness;
and \textit{3)} Using the extractor $\mathbf{E}(\cdot)$ with $\mathcal{L}_{2}$ distance achieves the best performance.

\noindent\textbf{EB Block Settings.} Firstly, we present PCC@M (\cref{fig:nmk}(a)), MSE (\cref{fig:nmk}(b)), and MAE (\cref{fig:nmk}(c)), by varying the number of exemplars used in the EB block from one to fifteen. Having nine exemplars find the best PCC@M and MAE, and we have the best MSE by using ten exemplars. Again, our task emphasizes capturing relative changes. Thus, our final recommendation is to use nine exemplars. Secondly, we ablate the EB block architectures and the frequency of interleaving with the vision transformer blocks (\cref{tab:blc}). This balances the model's capability in structuring the ViT representations and receiving knowledge from the exemplars. We have the best PCC@M, by setting heads, head dimension, and frequency to 8, 64, and 2.

\begin{table}[!t]
    \centering
    \vspace{-1em}
    \caption{Ablation study on the EB block.}
    \vspace{-0.7em}
    \resizebox{0.85\linewidth}{!}{
    \begin{tabular}{ccccc}
        \toprule
        {Heads} & {\makecell[l]{Head Dim}} & {Frequency} & {PCC@M$_{\times10^{1}}$}\\
        \midrule
         4  & 32 & 2 & 1.84\\
         4  & 64 & 2 & 1.76\\
         4  & 64 & 3 & 1.78\\
         8  & 32 & 2 & 1.85\\
         8  & 64 & 2 & \textbf{2.02}\\
         8  & 64 & 3 & 1.88\\
         16 & 32 & 2 & 1.86\\ 
         16 & 64 & 2 & 1.82\\ 
         16 & 64 & 3 & 1.80\\ 
        \bottomrule
    \end{tabular}}
    \label{tab:blc}
\end{table}

\noindent\textbf{Model Architectures.} We present the performance of `Pretrained $\mathbf{E}(\cdot)$', `Backbone only', `w/o EB block', and `w/o projector' in \cref{tab:net}.  `Pretrained $\mathbf{E}(\cdot)$' adds one trainable linear layer to $\mathbf{E}(\cdot)$, while freezing weights of $\mathbf{E}(\cdot)$ for gene expression prediction. This is also known as linear probing in the literature. `Backbone only' uses the ViT backbone architecture. Note that, we use a non-regular ViT architecture which is determined by a binary search. `w/o EB block' is equivalent to concatenating the pooled ViT representations with the global view for predicting gene expression. `w/o projector' replaces the projector with a linear layer to unify the dimension. 
Our findings are as follows:
\textit{1)} The `Pretrained $\mathbf{E}(\cdot)$' setting is a strong baseline. Note that, this setting has better performance than the STNet and NSL (\cref{tab:eva}). Our $\mathbf{E}(\cdot)$ captures gene expression-related features in an unsupervised manner, showing the potential of our unsupervised exemplar retrieval;
\textit{2)} The `w/o EB block' setting achieves worse performance than the `Backbone only' setting. Concatenating the global view for gene expression prediction with a single linear layer disables interactions between the global view and the ViT representations. This potentially causes inconsistent behaviors and results in poor performance; 
and \textit{3)} With all proposed components, we have the best performance. 

\begin{table}[!t]
    \centering
    \vspace{-1em}
    \caption{Ablation study on model architectures. 
    }
    \vspace{-0.7em}
    \resizebox{0.85\linewidth}{!}{
    \begin{tabular}{lccc}
        \toprule
        Settings & MSE$_{\times10^{2}}$ & MAE$_{\times10^{1}}$ & PCC@M$_{\times10^{1}}$  \\
        \midrule
        Pretrained $\mathbf{E}(\cdot)$ & 4.85 & 1.76 & 1.56 \\
        Backbone only & 4.31 & 1.67 & 1.80\\
        w/o EB Block &  4.65 & 1.72 & 1.70\\
        w/o projector & 4.22 & 1.63 & 1.85\\
        \midrule
        Ours & \textbf{4.10} & \textbf{1.55} & \textbf{2.02} \\
        \bottomrule
    \end{tabular}}
    \label{tab:net}
\end{table}

\section{Conclusion}
This paper proposes an \name~framework to accurately predict gene expression from each fine-grained area of tissue slide image, i.e., different windows. \name~uses the ViT as a backbone while integrating with exemplar learning concepts. We first have an extractor to retrieve the exemplars of the given tissue slide image window in an unsupervised manner. Then, we propose an EB block to progressively revise the ViT representation by reciprocating with the nearest exemplars. With extensive experiments, we demonstrate the superiority of the \name~framework over the SOTA methods. \name~is promising to facilitate studies on diseases and novel treatments with accurate gene expression prediction.

{\small
\bibliographystyle{ieee_fullname}
\bibliography{egbib}
}

\end{document}